\documentclass[runningheads]{llncs}

\usepackage{times}
\usepackage{epsfig}
\usepackage{graphicx}
\usepackage{amsmath}
\usepackage{amssymb}
\usepackage{multirow}
\usepackage{listings}
\usepackage{xcolor}
\usepackage[numbers]{natbib}

\usepackage{booktabs}
\usepackage{color}
\usepackage{floatrow}
\newfloatcommand{capbtabbox}{table}[][\FBwidth]

\usepackage[width=122mm,left=12mm,paperwidth=146mm,height=193mm,top=12mm,paperheight=217mm]{geometry}

\usepackage[breaklinks=true,bookmarks=false]{hyperref}

\newcommand{\tsx}{\textsuperscript}

\begin{document}

\pagestyle{headings}
\mainmatter

\title{Human Pose Estimation in Space and Time\\using 3D CNN}

\author{Agne Grinciunaite\inst{1,2} 
\and Amogh Gudi\inst{1}
\and Emrah Tasli\inst{1,3} 
\and Marten den Uyl\inst{1}}

\institute{VicarVision, Amsterdam, The Netherlands\\
\email{a.grinciunaite@gmail.com}
\and
Vilniaus Gedimino Technikos Universitetas, Vilnius, Lithuania
\and
Booking.com, Amsterdam, The Netherlands
}

\maketitle

\begin{abstract}
This paper explores the capabilities of convolutional neural networks to deal with a task that is easily manageable for humans: perceiving 3D pose of a human body from varying angles. However, in our approach, we are restricted to using a monocular vision system. For this purpose, we apply a convolutional neural network approach on RGB videos and extend it to three dimensional convolutions. This is done via encoding the time dimension in videos as the 3\tsx{rd} dimension in convolutional space, and directly regressing to human body joint positions in 3D coordinate space. 
This research shows the ability of such a network to achieve state-of-the-art performance on the selected Human3.6M dataset, thus demonstrating the possibility of successfully representing temporal data with an additional dimension in the convolutional operation.
\end{abstract}
	
\section{Introduction}
From a psychological stand point, it has been argued that humans detect real-world structures by detecting changes along physical dimensions (contrast values) and representing these changes (with respect to time) as relations (differences) along subjective dimensions \cite{Jones76time}. 
More directly, it has been suggested that the temporal dimension is necessary and is coupled with spatial dimensions in human mental representations of the world \cite{freyd1987dynamic}.
This implies merit in incorporating time into a definition of structure from a computer vision modelling point of view. This forms the inspiration for this work.
	
This work deals with a long-standing task in computer vision - human pose modelling in 3D from monocular videos. The challenges of this task include large variability in poses, movements, appearance and background, occlusions and changes in illumination. 
	
This paper proposes a method to estimate the body pose of a human (in terms of body joint locations in 3D) from video capture using a single 2D monocular camera via a deep three dimensional convolutional neural network. The key idea behind this approach is that time, as a dimension, could be encoded as the $Z$-dimension of 3D convolutional operation (where the other two $X$ and $Y$ dimensions are along the height and width of the image).  The hypothesis behind this is that temporal information can be efficiently represented as an additional dimension in deep convolutional neural networks (see \cite{tran2015learning, agne2016development} for a detailed description of 3D convolution).
It is important to note here that no depth information is provided to the network as input, and the system is expected to infer the location of body joint positions in all three spatial dimensions only based on the stream of 2D frames in the video.
A more detailed and complete description of this work can be found in \cite{agne2016development}.
	
Such a system can have applications in areas such as visual surveillance, human action prediction, emotional state recognition, human-computer interfaces, video coding, ergonomics, video indexing and retrieval, etc.
	
\section{Related Work} \label{sec:relatedwork}
There have been a number of studies carried out in the human pose estimation field using different generative and discriminative approaches. Most of the published works deal with still single \cite{wang2014robust} or depth images \cite{oberweger2015hands}. Also, most often it is attempting to estimate 2D full \cite{du2014full}, upper body \cite{ToshevS13} or single \cite{FanZLW15} joint position in the image plane. Additionally, many approaches incorporate 2D pose estimations or features to retrieve 3D poses \cite{zhou2016spatio, zhou2015sparseness}.
The work in \cite{ToshevS13} formulates 2D pose estimation as a joint regression problem, using a conventional deep CNN architecture. The predictions are further iteratively refined by analysing relevant regions within the images in higher resolution.
\cite{TompsonGJLB14} introduces a heat-map based approach, where a spatial pyramid input is used to generate a heat map describing the spatial likelihood of joint positions. 
\cite{pfister2014deep} presents an architecture similar to \cite{ToshevS13}, with a key difference being that multiple consecutive video frames are encoded as separate colour channels in the input. Although this approach appears similar to that of 3D CNNs, the key difference here is that this approach enforces the $Z$ dimension of the `3D' kernel to be equal to the number of channels. Therefore, the kernel has no space to convolve in this dimension.
The first architecture utilizing 3D CNNs was proposed in 2013 and applied to human action recognition in \cite{Ji3dconv}. As in our proposed work, the third spatial dimension of the convolution operation is used to encode the time dimension on the video stream. This work also utilizes recurrent neural networks to finally predict the human action category. However, they do not explore the use of 3D CNNs for predicting the precise locations of body joints. 
Recent methods tested on the Human3.6M dataset include a discriminative approach to 3D human pose estimation using spatiotemporal features (HOG-KDE) \cite{tekin2015predicting}, as well as a 2D CNN based 3D pose estimation framework (2DCNN-EM) \cite{zhou2015sparseness}. However, one of the drawbacks of these approaches is that they utilize a large number of frames in a sequence comparing to our proposed 3D CNN method.
	
Our approach studies the suitability of using 3D convolutional networks for the task of 3D pose estimation from 2D videos. To the extent of our knowledge, this is the first work to do so. More fundamentally, this work explores the effects of processing spatio-temporal data using three dimensional convolutions, where the temporal dimension in data is represented as a additional dimension in convolutions.

\section{Dataset}
Human3.6M Dataset \cite{h36m_pami} is so far the largest publicly available motion capture dataset. It consists of high resolution 50Hz video sequences from 4 calibrated cameras capturing 10 subjects performing 15 different actions (`eating', `posing', etc.). 3D ground truth joint locations as well as bounding boxes of human bodies are provided. 
Note that we consider videos from the 4 camera positions independently, and do not combine them in any way.
Our evaluation was done on 17 core joints from the available 32 joint locations. 
For official testing, the ground truth data for 3 subjects is withheld and used for results evaluation on the server. 

\section{Method}
\subsection{Pre-processing} \label{sec:preprocessing}
\begin{figure}
	\centering
	\includegraphics[width=0.8\linewidth]{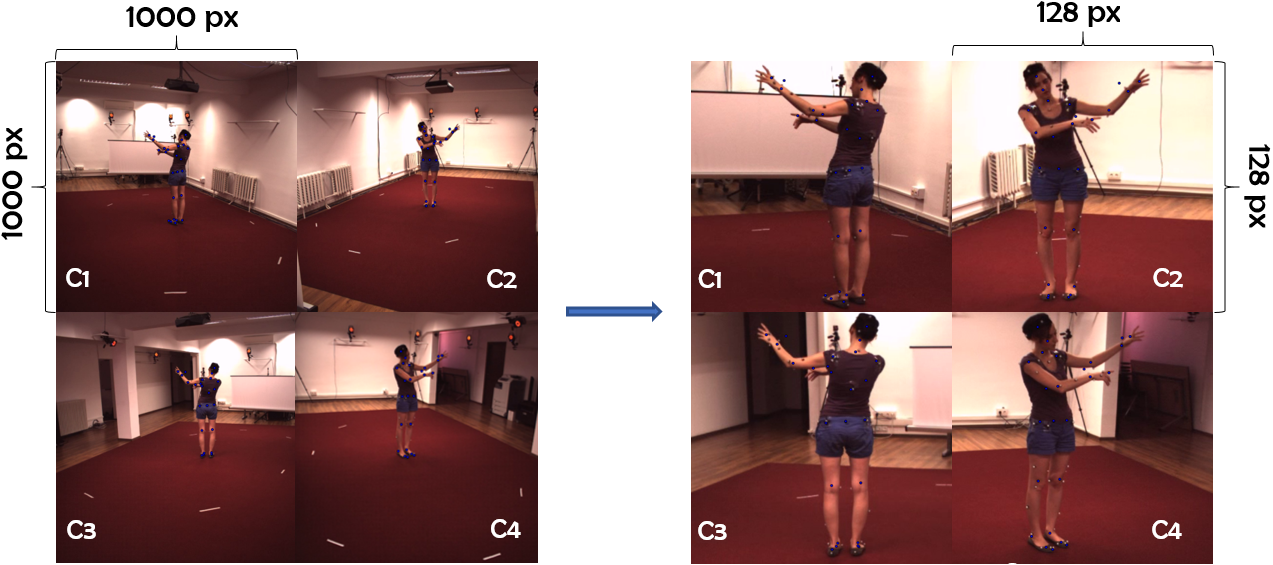}
	\caption[Image preprocessing]{Image pre-processing from 4 camera views capturing subject no. 1 performing action `Directions'}
	\label{fig:cropping}
\end{figure}

The original Human3.6M video frames are cropped using bounding box binary masks and extended to the larger side to make the crop squared. 
Cropped images are resized to 128$\times$128 resolution (chosen arbitrarily). The results of cropping can be seen in Figure \ref{fig:cropping}.

\paragraph{Data sampling}
Due to the large amount of available data, limited memory and time constrains, data sub-sampling is performed. One training data sample is composed of 5 sequential colour images with resolution of 128$\times$128. These were sampled from the original video to obtain a frame-rate of 13Hz. Random selection was performed from every chosen training, validation and testing subjects' videos to ensure that all the possible poses are selected.

\paragraph{Data alignment}
Ground truth joint positions were centered to the pelvis bone position (first joint).

\paragraph{Contrast normalization}
To reduce the variability that DNN needs to account for during training, global contrast normalization (GCN) was applied to the network's input data (per colour channel). 

\subsection{Deep 3D Convolutional Neural Network}
The final model of network's architecture was made up by starting with the small basic network with only three hidden 3D convolutional layers and building it up when testing with the small subset of data. Decisions on the construction parts and hyper-parameter selection were made by analysing experimental results and utilizing similar choices reported in related work reviewed in Section \ref{sec:relatedwork}. In this network, all the activations are PReLUs \cite{HeZR015} with \textit{p} set to 0.01.

The following equation provides a mathematical expression of discrete convolution (denoted by $*$) applied to three dimensional data ($\mathbf{X}$, of dimensions $m\times{n}\times{l}$), using three dimensional flipped kernels ($\mathbf{K}$):
\vspace{-5pt}
\begin{eqnarray}
(\mathbf{K} * \mathbf{X})_{i,j,k} = \sum_{m}\sum_{n}\sum_{l}\mathbf{X}_{i-m,j-n,k-l}\mathbf{K}_{m,n,l}
\end{eqnarray}
\vspace{-10pt}

\begin{figure*}
	\centering
	\includegraphics[width=\linewidth]{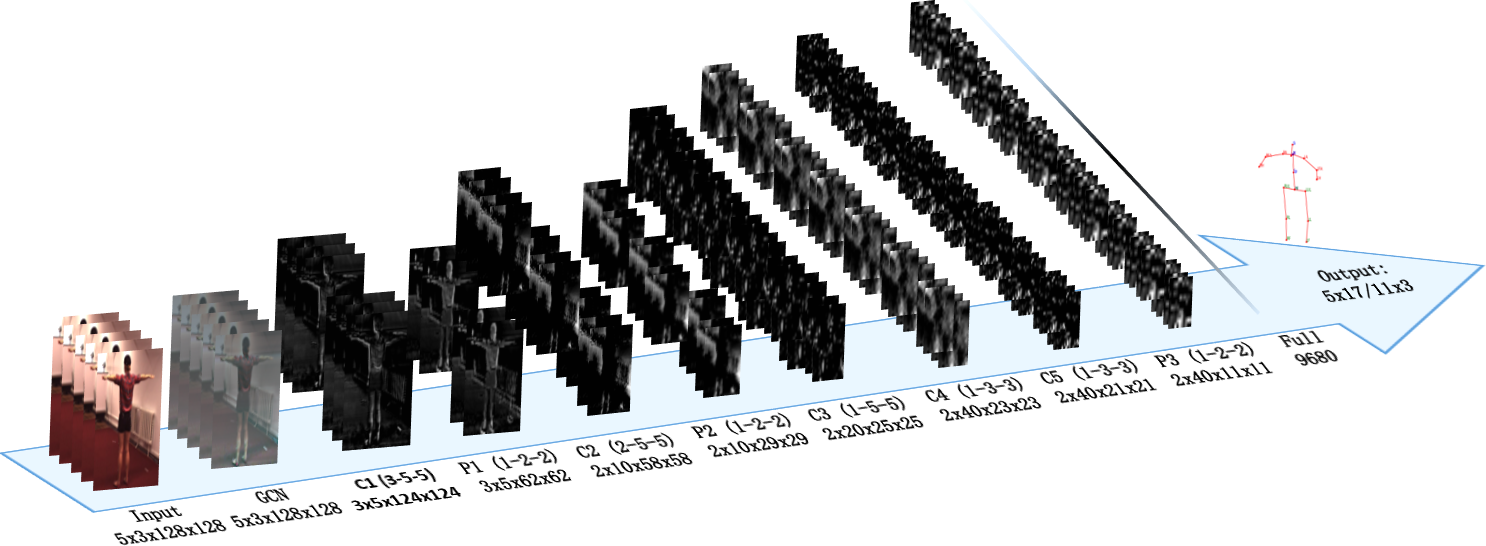}
	\caption[Proposed 3D CNN Architecture]{Proposed 3D CNN Architecture. Legend: C stands for convolutional layer, P for pooling layer; kernel sizes are specified in parenthesis; second row shows the size of corresponding layer's output; images show slices of some 3D activation maps per layer.}
	\label{fig:archi}
\end{figure*}

In our implementation, the stride is always equal to 1 and there is no zero-padding performed. Experiments have been completed with different kernel sizes and a number of convolutional layers in the network. The best performance was achieved with 5 convolutional layers with kernel sizes $3\times{5}\times{5}$, $2\times{5}\times{5}$, $1\times{5}\times{5}$, $1\times{3}\times{3}$ and $1\times{3}\times{3}$ respectively. Max pooling is performed after the first, second and fifth convolutional layers, and only on the image space with the kernel of size $2\times{2}$ (and not on the third time dimension).In our proposed architecture, the output of the last pooling layer is flattened to one dimensional vector of size 9680 and then is fully connected to the output layer of size 255 (5 frames $\times$ 17 joints $\times$ 3 dimensions). Complete 3D CNN architecture is shown in Figure \ref{fig:archi}. 

\paragraph{Training}
The network was trained using mini-batch (of size 10) stochastic gradient descent (with a learning rate of $10^{-5}$ and Nestrov momentum \cite{qian1999momentum} of $0.9$). Xavier initialization method \cite{glorot2010understanding} was used to set the initial weights, while the biases in convolutional layers were set to zero. 
Due to the memory and time  limitations, the maximum number of batches used was 20,000 for training, 2,000 for validation and 2,000 for testing (approximately half of the available data).
The cost function to be minimized during training was chosen to be the mean per joint position error (MPJPE)\cite{h36m_pami}, which is the mean euclidean distance between the true and predicted joint locations. This also serves as a good performance measure during testing.
Early stopping technique was used to avoid overfitting, where the training was terminated when the performance on the validation set stopped improving for 15 consecutive epochs.

\subsection{Post-Processing}
The shape of the network output contains estimated 3D joint positions for 5 consecutive frames. During inference time, this makes it possible to feed each video frames 5 times through the network at 5 different positions in the input sequence. This gives us 5 outputs for each frame. In order to get a more robust estimation, these overlapping outputs are averaged together.

\section{Results}
\begin{figure}
\begin{floatrow}
\ffigbox{
	\fbox{\includegraphics[width=0.83\linewidth]{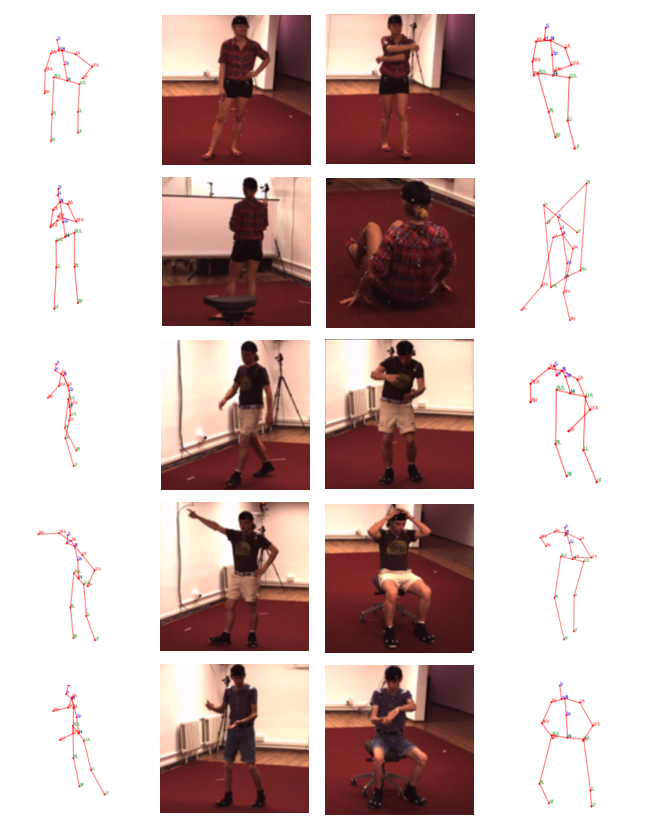}}
}{
	\caption[Selected results visualization]{Visualization of some 3D pose estimation results.}
	\label{fig:results}
}
\capbtabbox{%
	\centering
	\resizebox{\linewidth}{!}{
		\begin{tabular}{lccc}
			\toprule
		    \textbf{Action Subset} & \textbf{KDE \cite{h36m_pami}} & \textbf{3DCNN} & \textbf{\% Improvement}		                  \\
		         & (s.o.t.a)						   & (ours)			& (ours over s.o.t.a)			   	              \\
			\midrule                                                                                                                  
			Directions             & 117                           & \textbf{91}    & {\color[HTML]{009901}$\blacktriangle$      22\%} \\
			Discussion             & 108                           & \textbf{89}    & {\color[HTML]{009901}$\blacktriangle$      18\%} \\
			Eating                 & \textbf{91}                   & 94             & {\color[HTML]{FE0000}$\blacktriangledown$ \ -3\%} \\
			Greeting               & 129                           & \textbf{102}   & {\color[HTML]{009901}$\blacktriangle$      21\%} \\
			Phoning                & \textbf{104}                  & 105            & {\color[HTML]{FE0000}$\blacktriangledown$ \ -1\%} \\
			Posing                 & 130                           & \textbf{99}    & {\color[HTML]{009901}$\blacktriangle$      24\%} \\
			Purchases              & 134                           & \textbf{112}   & {\color[HTML]{009901}$\blacktriangle$      16\%} \\
			Sitting                & \textbf{135}                  & 151            & {\color[HTML]{FE0000}$\blacktriangledown$ -12\%} \\
			Sitting Down           & \textbf{200}                  & 239            & {\color[HTML]{FE0000}$\blacktriangledown$ -20\%} \\
			Smoking                & 117                           & \textbf{109}   & {\color[HTML]{009901}$\blacktriangle$     \ 7\%} \\
			Taking Photo           & 195                           & \textbf{151}   & {\color[HTML]{009901}$\blacktriangle$      23\%} \\
			Waiting                & 132                           & \textbf{106}   & {\color[HTML]{009901}$\blacktriangle$      20\%} \\
			Walking                & 115                           & \textbf{101}   & {\color[HTML]{009901}$\blacktriangle$      12\%} \\
			Walking with Dog       & 162                           & \textbf{141}   & {\color[HTML]{009901}$\blacktriangle$      13\%} \\
			Walking Together       & 156                           & \textbf{106}   & {\color[HTML]{009901}$\blacktriangle$      32\%} \\
			\midrule                                                                                                                  
			\textbf{Average}       & 133                           & \textbf{119}   & {\color[HTML]{009901}$\blacktriangle$      11\%} \\
			\bottomrule
		\end{tabular}
	}
}
{%
\caption{Results comparing with the state-of-the-art (s.o.t.a) on the Human3.6M test set. Legend: Numbers denote MPJPE error in $mm$ (less is better).}
\label{tab:bestresults}
}
\end{floatrow}
\end{figure}

In Table \ref{tab:bestresults} the best results are compared with state-of-the-art reported on the dataset website. All the numbers are MPJPEs in millimetres. It can be seen that network performs better on 11 actions and the MPJPE is 11\% smaller on average. However, the model performs worse on the actions where people are sitting on the chair or on the ground showing difficulties to deal with body part occlusions. Figure \ref{fig:results} shows some selected examples of pose estimation by the network. 
This could also be due to the fact that the temporal window of 5 frames is too short to capture these joint positions.
Expanding the window or incorporating recurrent neural networks in this architecture could handle this better by capturing longer-term trajectories.

On further investigation, it was also found that the joint position of freely moving upper body joints like hands were relatively poorly predicted. Countering this, a further improvement in performance was obtained by training a separate network to estimate only the upper body joints, and merging the outputs together.

Unfortunately, the two most recent works in 3D pose estimation on the Human3.6M dataset by \cite{tekin2015predicting, zhou2015sparseness} fail to report their scores on the official test sets, thereby making it very hard to compare out works. However, they do report average MPJPE scores of 124 (\cite{zhou2015sparseness}) and 113 (\cite{tekin2015predicting}) on two male subjects (S9 and S11, which are in our training set). 

Additionally, a comparison was performed with a 2D convolution based model with an otherwise identical architecture and training. It was found that our 3D CNN architecture outperforms this 2D CNN based network even without the post-processing step, thereby suggesting that modelling temporal dynamics improves 3D human pose estimation, perhaps due to inherent body-joint trajectory tracking.

The average processing time per 5-frame sample during testing was about 1ms / 13ms on a Nvidia GTX 1080 GPU / Intel Xeon E5 CPU, implying real-time frame rates.

\section{Conclusions}
A discriminative 3D CNN model was implemented for the task of human pose estimation in 3D coordinate space using 2D RGB video data. To the best of our knowledge, this is the first attempt to utilize 3D convolutions for the formulated task. 
It was shown that such a model can cope with 3D human pose estimation in videos and outperform the existing methods on the Human3.6M dataset. Proposed model was officially tested on dataset provider's evaluation server and compared with other reported results, which it could outperform with real-time processing speeds. 
These results suggest that time can be successfully encoded as an additional convolutional dimension for the task of modelling real world objects from 2D sequence of images.
\paragraph{Future Work}
There are a number of possible future work directions that can extend this work: More hyper-parameter tuning and utilizing higher computational resources could possibly lead to more accurate estimations; testing model's capabilities on other available datasets; expanding the temporal window and/or combining the proposed model with recurrent neural networks (known for their ability to process temporal information).

\clearpage
\bibliographystyle{splncs03}
\bibliography{MyBib}

\end{document}